\documentclass[10pt,journal,compsoc]{IEEEtran}
\usepackage{amsmath,amssymb,amsfonts}
\usepackage{graphicx}
\usepackage{textcomp}
\usepackage{multirow}
\usepackage{algpseudocode}
\usepackage{caption}
\usepackage{subcaption}
\usepackage{url}

\ifCLASSOPTIONcompsoc
  \usepackage[nocompress]{cite}
\else
  \usepackage{cite}
\fi
\ifCLASSINFOpdf
\else
\fi
\begin{document}
%
\title{ChatGPT Encounters Morphing Attack Detection: Zero-Shot MAD with Multi-Modal Large Language Models and General Vision Models}
%
%
%
%

\author{Haoyu Zhang$^\dag$, ~\IEEEmembership{Member,~IEEE,} ~Raghavendra Ramachandra$^\dag$, ~\IEEEmembership{Senior Member,~IEEE,}\\ ~Kiran Raja$^\dag$, ~\IEEEmembership{Senior Member,~IEEE,}   ~Christoph Busch $^\dag$, ~\IEEEmembership{Senior Member,~IEEE,}$^\ddagger$ \\
$^\dag$ Norwegian University of Science and Technology (NTNU), Gjøvik, Norway\\
$^\ddagger$ Darmstadt University of Applied Sciences (HDA), Darmstadt, Germany\\
Email: \{haoyu.zhang, raghavendra.ramachandra, kiran.raja, christoph.busch\}@ntnu.no\\ \thanks{Haoyu Zhang and Raghavendra Ramachandra contributed equally to this work.}
}

\IEEEtitleabstractindextext{%
\begin{abstract}
Face Recognition Systems (FRS) are increasingly vulnerable to face-morphing attacks, prompting the development of Morphing Attack Detection (MAD) algorithms. However, a key challenge in MAD lies in its limited generalizability to unseen data and its lack of explainability—critical for practical application environments such as enrolment stations and automated border control systems. Recognizing that most existing MAD algorithms rely on supervised learning paradigms, this work explores a novel approach to MAD using zero-shot learning leveraged on Large Language Models (LLMs). 
We propose two types of zero-shot MAD algorithms: one leveraging general vision models and the other utilizing multimodal LLMs. For general vision models, we address the MAD task by computing the mean support embedding of an independent support set without using morphed images. For the LLM-based approach, we employ the state-of-the-art GPT-4 Turbo API with carefully crafted prompts.
To evaluate the feasibility of zero-shot MAD and the effectiveness of the proposed methods, we constructed a print-scan morph dataset featuring various unseen morphing algorithms, simulating challenging real-world application scenarios. Experimental results demonstrated notable detection accuracy, validating the applicability of zero-shot learning for MAD tasks. Additionally, our investigation into LLM-based MAD revealed that multimodal LLMs, such as ChatGPT, exhibit remarkable generalizability to untrained MAD tasks. Furthermore, they possess a unique ability to provide explanations and guidance, which can enhance transparency and usability for end-users in practical applications.
\end{abstract}
 }


\maketitle

\IEEEdisplaynontitleabstractindextext

%
\IEEEpeerreviewmaketitle

\IEEEraisesectionheading{\section{Introduction}\label{sec:intro}}

%
%
%
%

\IEEEPARstart{F}{ace} Recognition Systems (FRS) have found widespread use in securing access control for various applications \cite{jain2011handbook}. However, the improvements in generalizability and robustness against noise and image degradation have, in turn, increased these systems’ vulnerability to sophisticated attacks \cite{ngan2021face}. One such threat is the face-morphing attack, where images from multiple individuals are combined into a single, manipulated face image \cite{ferrara2014magic}.
In many countries, passport applications still rely on printed photographs and paper forms \cite{venkatesh2020detecting}. This traditional approach presents an opportunity for malicious actors: by working with an accomplice, they can submit carefully altered, morphed images during the enrolment stage. If a travel document containing a morphed photograph is successfully registered, the attacker could subsequently use it to pass through border controls undetected, effectively assuming the accomplice’s identity.

An input morphing image often contains visible artifacts or unnatural traces arising from the morphing process, unless it undergoes thorough post-processing \cite{UBO_Morphing_Tool}  \cite{PostProcessMorph} \cite{zhang2021mipgan} \cite{zhang2023morph}. Additionally, subtle details or concealed indications of morphing may be difficult for human observers to detect \cite{godage2022analyzing}. To address these challenges, researchers have focused on developing Single-Image-Based Morphing Attack Detection (S-MAD) algorithms. Compared to Differential-Image-Based Morphing Attack Detection (D-MAD) approaches, S-MAD is inherently more difficult because it lacks a trusted pristine reference image. Moreover, since S-MAD relies on a single input image rather than an image pair, it cannot leverage sample combinations to increase the scale and diversity of the training data. This limited data availability makes it challenging to achieve highly accurate S-MAD algorithms \cite{9380153}.


Deep learning techniques have made substantial progress in improving the generalization of computer vision tasks. Researchers have applied these methods to MAD as well \cite{9380153}. However, due to privacy regulations, collecting high-quality, large-scale face morphing datasets remains challenging, making it difficult to develop well-generalized S-MAD algorithms using state-of-the-art deep learning models. Moreover, generating print-scanned data to simulate real-world application scenarios is both time-consuming and resource-intensive. Despite these challenges, ensuring strong generalization is critical for MAD in real-life scenarios. First, real-world morphing attacks are essentially open-set problems, requiring MAD algorithms to detect morphs generated by previously unseen morphing algorithms. Second, in common scenarios such as enrolment stations, the algorithm must handle print-scanned images, where image quality is limited and morphing traces are more subtle. 

\begin{figure*}[ht!]
\centering
   \includegraphics[width=0.95\linewidth]{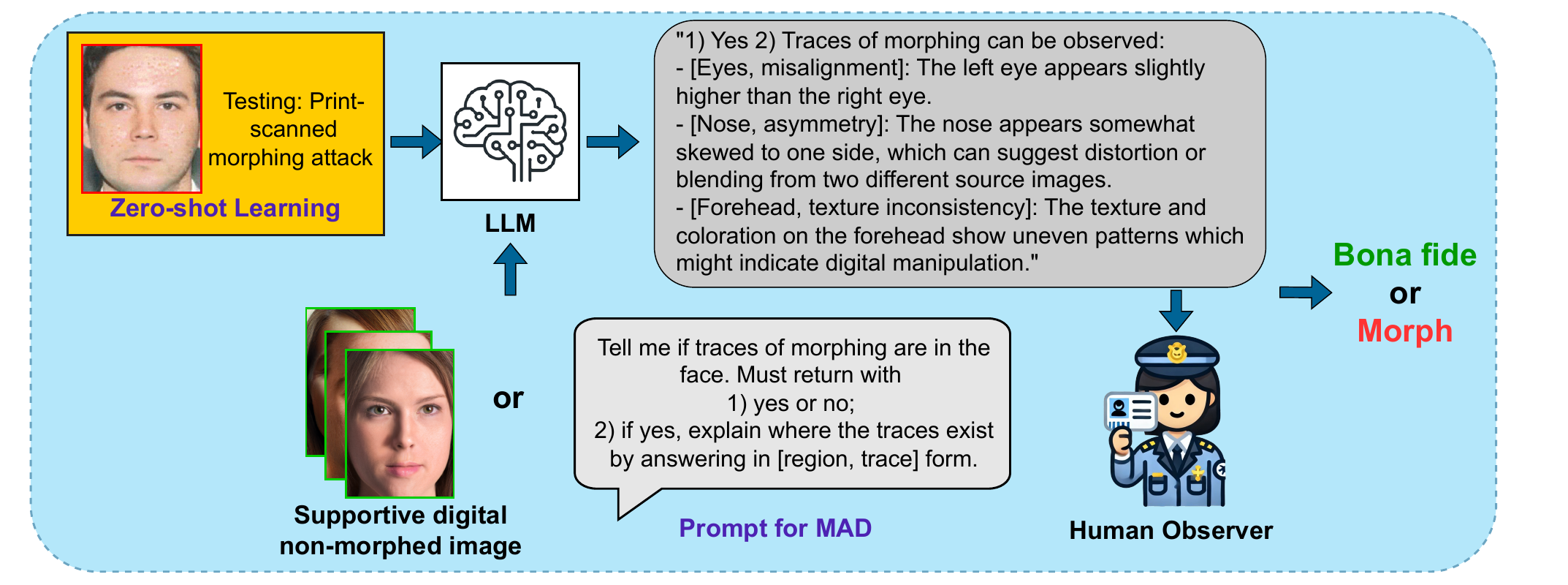}
   \caption{Overview of the proposed method using large language model for zero-shot MAD.}
\label{fig:overview}
\end{figure*}

By reinterpreting the S-MAD task as a classification problem, we note that existing solutions predominantly rely on supervised learning approaches, which often struggle to generalize to unseen attacks in real-world scenarios. Moreover, these methods frequently lack interpretability, making it difficult to explain results to end-users \cite{9380153}. Motivated by these challenges, we introduce a zero-shot learning approach for S-MAD that utilizes a Large Language Model (LLM), as illustrated in Figure \ref{fig:overview}. In this zero-shot setting, print-scanned non-morphed and morphed face images are considered unseen classes during the training phase. We employ GPT4 Turbo \cite{achiam2023gpt}, a state-of-the-art LLM model, and design various prompts tailored to the MAD task, enabling effective classification of both print-scanned morphs and bona fide face images. Our key contributions are summarized as follows\footnote{Scripts will be released at: \\ \url{https://share.nbl.nislab.no/HaoyuZhang/smad_gpt}}:
\begin{itemize}
    \item We propose a zero-shot single-image-based morphing attack detection (ZS-MAD) method that leverages a state-of-the-art LLM and explores various prompt strategies. The key contribution of this work lies in designing prompts that ensure reliable morphing attack detection. To the best of our knowledge, this is the first study to investigate the use of LLMs for face morphing detection in a zero-shot learning context. 
    \item During experiments, we developed an evaluation protocol using different morphing algorithms and evaluated the detection accuracy, stability, failure rate, and interpretability of LLM for MAD.
    \item To evaluate the MAD performance of the LLM-based ZS-MAD algorithm, we propose a comparative method utilizing general vision models.
\end{itemize}

The rest of the paper is organised as follows: Firstly in Section \ref{sec:related}, we motivate our work by reviewing existing relevant studies about using LLM for biometrics, and then connecting to the challenges and our objectives for S-MAD task. Then in Section \ref{sec:proposed}, the proposed method of ZS-MAD using LLM and vision models is described in detail. Further in Section \ref{sec:exp}, we include our experiment settings and results to benchmark the proposed ZS-MAD methods using LLM and general vision models and also conduct a comprehensive analysis on the applicability of using LLM for MAD. Discussions based on the obtained results and observations are included in the following Section \ref{sec:discussion}. Finally, we conclude the paper and provide insights into future work in Section \ref{sec:conclusion}.

\section{Related Works}
\label{sec:related}
Several studies have explored the use of Large Language Models (LLMs) in biometrics and security applications. Most of these efforts focus on evaluating the performance of LLMs in recognition tasks, including face recognition \cite{hassanpour2024chatgpt} \cite{deandres2024good}, soft-biometric estimation \cite{hassanpour2024chatgpt} \cite{deandres2024good}, iris recognition \cite{farmanifard2024chatgpt}, and gait recognition \cite{chivereanu2024aligning}. Overall, LLMs have demonstrated considerable performance and strong generalization capabilities in these tasks, which primarily rely on visual appearances.

However, face morphing attack detection mainly focuses on detecting minor traces distributed among the face region and is hence different compared to general biometric tasks \cite{venkatesh2021face}. Recent works have also shown the applicability of LLM for digital forgery detection and face anti-spoofing \cite{wu2023cheap} \cite{jia2024can} \cite{shi2024shield} which is closer to the S-MAD applications. Wu et al. \cite{wu2023cheap} illustrate that LLM is capable of understanding the concept of face morphing given a set of gradually morphed faces. However, the detailed evaluation of S-MAD and the study on S-MAD with print-scanned data were not covered in these works. 

Single-image-based morphing attack detection (S-MAD) algorithms can generally be categorized into two types: explicit and implicit. Explicit algorithms \cite{batskos2023visualizing} \cite{singh2022fusion} \cite{raghavendra2022multimodality} \cite{Debiasi-PRNUVarianceMAD-BTAS-2018} rely on handcrafted features and traditional machine learning models, offering better interpretability. In contrast, implicit algorithms \cite{borghi2023revelio} \cite{zhang2024generalized} use more powerful but less transparent approaches, such as deep learning networks, which indicates a trade-off between generalizability and explainability. When it comes to generalization, both explicit and implicit methods often depend on task-specific training and face limitations due to the scarcity of large-scale, high-quality datasets—a consequence of privacy regulations. Regarding explainability, current solutions are either imprecise with gradient-based techniques \cite{seibold2021focused} \cite{myhrvold2022explainable} or remain inaccessible to end-users lacking technical expertise. Recently, Patwardhan et al. \cite{patwardhanempowering} explored leveraging the CLIP model with image-text inputs to achieve more interpretable MAD outcomes. These challenges motivate us to explore LLM-based, zero-shot learning approaches that can enhance both the generalizability and explainability of MAD without the need for extensive, task-specific training data.

\section{Proposed Method}
\label{sec:proposed}
In this section, we describe our two proposed methods for zero-shot MAD. The first method uses multi-modal LLM with different prompts designed for the MAD task. These prompts are designed for both utility and also to have a further comprehensive study on the applicability of using LLM for MAD. Afterwards, to compare with the LLM-based method, another zero-shot MAD method using general vision models such as CNNs pre-trained on image classification tasks is introduced.  
\subsection{Zero-Shot MAD based on Multi-Modal LLM}
For zero-shot MAD using a multimodal LLM, we apply the state-of-the-art GPT4-Turbo (gpt-4-turbo-2024-04-09) API and carefully craft various prompts for the S-MAD task. The model is neither specifically trained for MAD tasks nor familiar with print-scanned images. We provide the model with both a suspicious input image and a carefully designed prompt, enabling it to produce text-based responses. Since the prompts contain semantic information to guide the model in identifying previously unseen classes, different prompt strategies can yield varying results. Inspired by Jia et al. \cite{jia2024can} and the Chain-of-Thought (CoT) approach \cite{chivereanu2024aligning}, we design our prompts to include two key components: a CoT element that allows the LLM to reason through the problem, and a questioning prompt that frames the MAD task itself. Because MAD is a zero-shot task for the LLM, there is a risk that the model may refuse to perform the required analysis, considering it beyond its capabilities. For instance, Jia et al. \cite{jia2024can} have reported declined cases from LLM for deepfake detection task due to a lack of context or safety
concerns. One can try the Gemini 1.5 Pro API for MAD task\footnote{https://deepmind.google/technologies/gemini/pro/} which often declined these requests. To overcome this issue, we leverage the CoT component and the GPT4-Turbo API. Below, we describe the CoT component and the various questioning prompts we designed to facilitate effective zero-shot MAD detection.

\begin{itemize}
    \item  \textbf{Prompt CoT}: I want you to act as an image forensic expert to check suspicious face images. Think step-by-step, first analyze the visual characteristics of the image and then answer.
    \item \textbf{Prompt 1}: Tell me if this is a face morphing attack. You must ONLY answer yes or no.
    \item \textbf{Prompt 2}: Tell me if this is a bona fide image. You must ONLY answer yes or no.
    \item \textbf{Prompt 3}: Tell me if this is a face morphing attack or a bona fide image. You must ONLY answer morphing attack or bona fide.
    \item \textbf{Prompt 4}: Tell me the probability of this image being a face morphing attack. You must ONLY answer a probability score between 0 and 100.
    \item \textbf{Prompt 5}: Tell me the probability of this image being a bona fide image. You must ONLY answer a probability score between 0 and 100.
    \item \textbf{Prompt 6}: Tell me the probability of this image being a face morphing attack or a bona fide image. You must ONLY answer a probability score between 0 (hundred-percent bonafide) and 100 (hundred-percent morphing attack).
    \item \textbf{Prompt 7}: Tell me if traces of morphing are in the face. Must return with 1) yes or no; 2) if yes, explain where the traces exist by answering in [region, trace] form.
    \item \textbf{Prompt 8}: Tell me if the image has the artifact attribute listed in the following list and return 1) yes or no; 2) if yes, the attribute number(s) if you have noticed in this image. The artifact list is [1- asymmetric eye iris; 2-strange artifacts around eye iris; 3-strange artifacts around nose; 4-strange artifacts around eyebrow; 5-irregular teeth shape or texture; 6-irregular ears or earrings; 7-strange hair texture; 8-inconsistent skin texture; 9-inconsistent lighting and shading; 10-strange background; 11-unnatural edges].
\end{itemize}
For Prompts 1 and 2, we use prompts to guide the LLM for single-class classification. As the prompt is single-classed, the answer of LLM may be biased to the question. For example, asking whether the image is a morph as Prompt 1 may increase the tendency of LLM to classify more images as morphing attacks. Hence, we include the corresponding Prompt 2 so that we can further study the bias caused by the prompt and whether the LLM is giving reasonable classification results. After the single-class classification, in Prompt 5 we study prompting with binary classification and let the model reason by itself without adding any restrictions on the output format. Further, we design Prompts 4-6, asking for a probability score which can be further used to conduct a comprehensive evaluation of the trade-off of the detection error rates, which is more common and meaningful for practical applications with adjustable thresholds. 

Building on the Chain-of-Thought (CoT) technique, research has demonstrated that encouraging the model to articulate its intermediate reasoning steps enhances performance on complex tasks \cite{wei2022chain}. In Prompt 7, we explicitly direct the model to identify suspicious regions as an explanation for its decision, thereby improving its reasoning capabilities. Finally, in Prompt 8, instead of having the model merely summarize and reason, we provide a list of common morphing artifacts—derived from human expert knowledge—and ask the model to select the traces it has observed. This approach serves as a bridge between expert insights and S-MAD algorithms, ensuring that the results are both explainable and easily understood through natural language descriptions.


Because LLM outputs can vary due to their inherent probabilistic nature, we conduct our evaluation in five rounds using the same input and prompts. We then average the results across these rounds, reducing the impact of randomness. This approach also allows us to perform a more comprehensive quantitative analysis of the stability and consistency of using LLMs for zero-shot MAD.

\subsection{Zero-Shot MAD based on General Vision Models}
As mentioned in Section \ref{sec:related}, current S-MAD approaches are mainly based on supervised task-specific training paradigms. To have a better comparison with the LLM as a zero-shot model, we propose another ZS-MAD method using general open-source vision models such as VGG \cite{simonyan2014very} and ResNet \cite{he2016deep} model pre-trained for the image classification task. For vision models, it is difficult to transfer the model pre-trained on other tasks to the S-MAD task with zero-shot learning, especially for the S-MAD task simulating real applications with unseen print-scanned data. Inspired by zero-shot image classification in multi-modality models \cite{radford2021learning}, we propose to use an additional digital bona fide to find a supportive anchor point. More specifically, given a pre-trained vision model $\mathcal{F}$, we first use a set of bona fide digital images $I'$ and compute the mean embedding extracted using the vision model:
\begin{equation}
    v_{anchor} = \frac{1}{N}\sum_{i=1}^{N} \mathcal{F}(I'_i).
\end{equation}
Then during inference, the extracted embedding of the print-scanned suspicious input face image $v_{in} = \mathcal{F}(I_{in})$  will be measured with the distance of the anchor embedding and the extracted input embedding. The classification score is calculate by: $d={distance}(v_{anchor},v_{in})$. We selected models trained on image classification with distances measured by Euclidean distance and cosine distance, which are commonly used in transfer learning of existing morphing attack detection methods \cite{raja2017transferable} \cite{scherhag2020deep}.

\section{Experiments and Results}
\label{sec:exp}
In this section, we first present the details of our constructed face morphing dataset in Section \ref{sec:exp_dataset}, followed by an introduction to our experimental settings and results. The initial set of experiments provides a benchmark of MAD accuracy for the proposed ZS-MAD approach under various prompt configurations for both LLM-based and general vision model-based methods, as described in Section \ref{sec:exp_mad}. We then further examine the consistency, stability, and explainability of using LLMs for S-MAD, offering a more comprehensive evaluation as a pilot study in this domain.

\subsection{Dataset}
\label{sec:exp_dataset}
To comprehensively assess the generalization capabilities across various morphing types, we selected three representative, state-of-the-art algorithms: the landmark-based LMA-UBO \cite{UBO_Morphing_Tool}, the GAN-based MIPGAN-II \cite{zhang2021mipgan}, and the diffusion-based Morph-PIPE \cite{zhang2023morph}. \textit{To comply with privacy regulations when evaluating the commercial LLM, we constructed our dataset using high-quality synthetic face images from the SynMorph dataset \cite{zhang2024synmorph}.} The non-morphed images were generated using a StyleGAN2 model \cite{karras2020analyzing} trained on the FFHQ dataset \cite{karras2019style}, from which we selected 50 male and 50 female synthetic subjects, resulting in 100 non-morphed samples. To create morph pairs, each selected subject was paired with another subject (not among the chosen 100) from the SynMorph dataset based on their facial similarity, as determined by the VGGFace2 \cite{simonyan2014very} model, ensuring no cross-gender pairs. As a result, our dataset consists of 100 non-morphed images and 300 morphed images, with each of the three morphing algorithms contributing 100 images.  A realistic application environment is simulated by printing and scanning all face images using a RICOH IMC6000 office printer at 1200 DPI for printing and 600 DPI for scanning. This setup adheres to ICAO standards \cite{ICAO-9303-p1-2021} and ensures the dataset is representative of real-world conditions. 
\begin{figure*}[h!]
    \centering
   \includegraphics[width=0.7\linewidth]{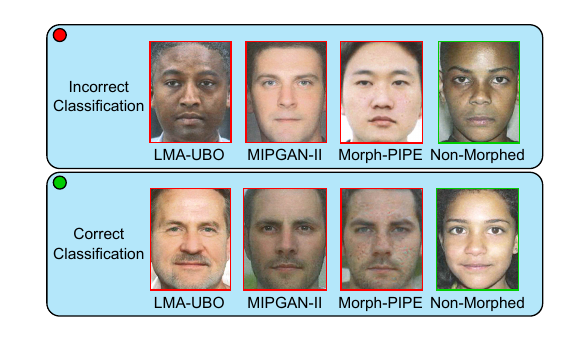}
   \caption{Example of extreme cases in Prompt 3 with incorrect and correct classification.}
\label{fig:example_prompt3}
\end{figure*}
Specifically for the ZS-MAD using vision models, an additional set of 50 non-morphed synthetic face images are randomly selected from the SynMorph dataset as the supportive set. It should be noted that only digital samples are used to extract and compute the anchor embedding, so that the print-scanned non-morphed data and print-scanned morphed data remain unseen classes following the zero-shot learning paradigm.  Finally, all of the images from our constructed dataset are detected and cropped by MTCNN model \cite{zhang2016joint} as our region of interest. we visualized our dataset by showing some extreme cases (predicted with a very high or very low score) of incorrect and correct classification results from Prompt 3 in Figure \ref{fig:example_prompt3}. It is shown that the morphed images generated by the selected algorithms are very realistic and challenging to determine between non-morphed samples, particularly after the print-scan effect that covers minor traces of morphing.

\subsection{MAD benchmark}
\label{sec:exp_mad}

The first part of our experiment involves MAD benchmark results across various configurations. To make the comparison more clear, we evaluate both different prompts for ZS-MAD using LLMs and assess different configurations for ZS-MAD using vision models, and the overall results. To measure MAD performance, we use standardized metrics: Morphing Attack Classification Error Rate (MACER) and Bona Fide Presentation Classification Error Rate (BPCER), as defined in ISO/IEC DIS 20059 \cite{ISO-IEC-20059}. MACER represents the proportion of morphing attacks misclassified as bona fide presentations, while BPCER represents the proportion of bona fide presentations misclassified as morphing attacks. We also employ Detection Error Trade-off (DET) curves \cite{martin1997det} to visualize the balance between MACER and BPCER. In addition, we report the Equal Error Rate (EER), where BPCER equals MACER, as a single scalar metric for easy comparison.

Different prompts produce a variety of response formats from the model, necessitating different approaches for evaluation. For Prompts 1, 2, 3, 7, and 8, the model’s output is essentially binary, so we map the predicted class directly to a binary label: 0 for non-morph and 1 for morph. This binary categorization simplifies the comparison of these prompts and ensures uniformity in the evaluation process. In contrast, Prompts 4, 5, and 6 produce continuous scores rather than explicit binary decisions. To incorporate these continuous scores into the benchmarking process, we rescale them to a normalized range of [0,1]. This rescaling ensures that both binary and continuous outputs can be meaningfully compared against the ground truth labels.
By applying these standardized labelling and rescaling strategies, we can fairly and consistently evaluate the MAD performance across all prompts.
\begin{figure*}[h!]
     \centering
     \begin{subfigure}[b]{0.32\textwidth}
         \centering
         \includegraphics[width=\textwidth]{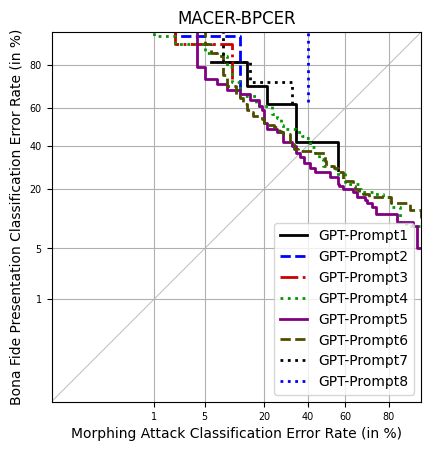}
         \caption{LMA-UBO}
         \label{fig:MAD_LLM_lma}
     \end{subfigure}
     \hfill
     \begin{subfigure}[b]{0.32\textwidth}
         \centering
         \includegraphics[width=\textwidth]{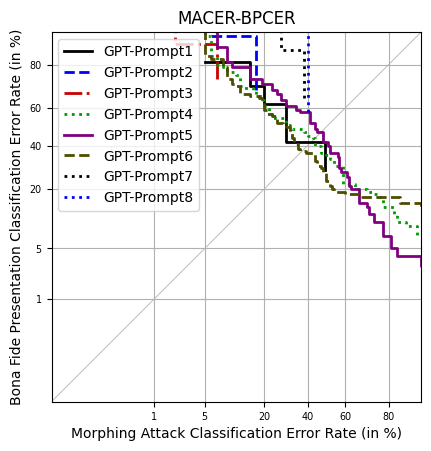}
         \caption{MIPGAN-II}
         \label{fig:MAD_LLM_mipgan}
     \end{subfigure}
     \hfill
     \begin{subfigure}[b]{0.32\textwidth}
         \centering
         \includegraphics[width=\textwidth]{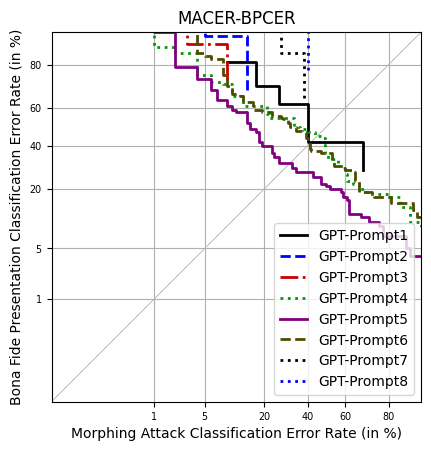}
         \caption{Morph-PIPE}
         \label{fig:MAD_LLM_pipe}
     \end{subfigure}
        \caption{DET plot of ZS-MAD using LLM with different prompts. }
        \label{fig:MAD_LLM}
\end{figure*}
Figure \ref{fig:MAD_LLM} shows the DET plots of ZS-MAD using LLM with different prompts. It is shown that the simple single or binary classification prompts lead to biased curves only appearing in the top-left region, indicating biased classification results. Though the LLM have considerable capability of detecting morphing attacks, most of the non-morphed images are misclassified as attacks. When comparing the curves with contrary prompts, differences can be noticed but the behavior is not contrary. This illustrates prompting indeed has an influence on the tendency of the model's MAD decision, while the results from LLM are still based on reasoning instead of random answers based on the prompt.

By further looking at the curves with continuous scores using prompts 4, 5, 6, it is shown that prompt 5 has shown the lowest equal error rate for LMA-UBO morphs in Figure \ref{fig:MAD_LLM_lma} and especially Morph-PIPE morphs in Figure \ref{fig:MAD_LLM_pipe}. However, the performance obviously decreases on MIPGAN-II morphs. This can be attributed to the domain gap between synthetic non-morphed data and bona fide images. It is aggravated for our testing protocol with MIPGAN-II morphs, where both non-morphed images and morphed images are generated by the StyleGAN2 \cite{karras2020analyzing} model.

\begin{figure*}[h!]
     \centering
     \begin{subfigure}[b]{0.32\textwidth}
         \centering
         \includegraphics[width=\textwidth]{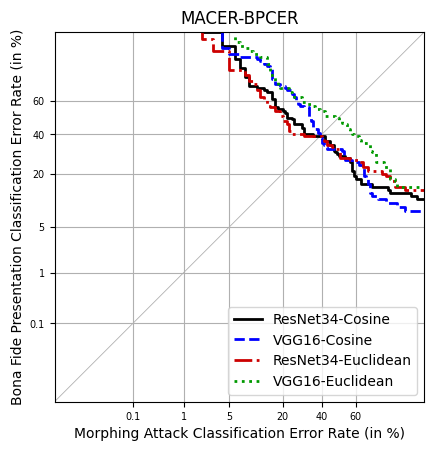}
         \caption{LMA-UBO}
         \label{fig:MAD_VM_lma}
     \end{subfigure}
     \hfill
     \begin{subfigure}[b]{0.32\textwidth}
         \centering
         \includegraphics[width=\textwidth]{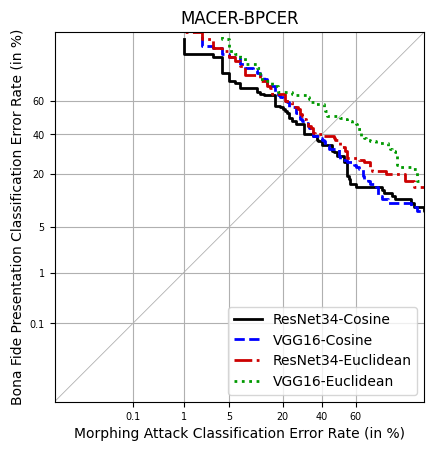}
         \caption{MIPGAN-II}
         \label{fig:MAD_VM_mipgan}
     \end{subfigure}
     \hfill
     \begin{subfigure}[b]{0.32\textwidth}
         \centering
         \includegraphics[width=\textwidth]{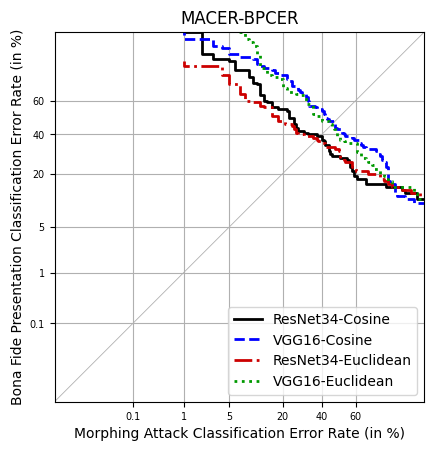}
         \caption{Morph-PIPE}
         \label{fig:MAD_VM_pipe}
     \end{subfigure}
        \caption{DET plot of ZS-MAD using vision models with different configurations.}
        \label{fig:MAD_VM}
\end{figure*}

\begin{table*}[h!]
\centering
\begin{tabular}{|cc|c|c|c|c|}
\hline
\multicolumn{2}{|c|}{}                                      & LMA-UBO & MIPGAN-II & Morph-PIPE & Overall (average) \\ \hline
\multicolumn{1}{|c|}{\multirow{2}{*}{ResNet34}} & Cosine    & 39.0   & \textbf{37.0}       & 39.0         & 38.3            \\ \cline{2-6} 
\multicolumn{1}{|c|}{}                          & Euclidean & 39.0   & 40.0     & 37.0         & 38.7       \\ \hline
\multicolumn{1}{|c|}{\multirow{2}{*}{VGG16}}    & Cosine    & 39.0   & 39.0     & 48.0      & 42.0       \\ \cline{2-6} 
\multicolumn{1}{|c|}{}                          & Euclidean & 50.0      & 51.0     & 46.0      & 49.0       \\ \hline
\multicolumn{2}{|c|}{GPT-Prompt4}                           & 40.0   & 44.0     & 45.0      & 43.0       \\ \hline
\multicolumn{2}{|c|}{GPT-Prompt5}                           & \textbf{36.0}     & 47.0       & \textbf{31.0}        & \textbf{38.0}               \\ \hline
\multicolumn{2}{|c|}{GPT-Prompt6}                           & 37.0     & \textbf{37.0}       & 41.0        & 38.3               \\ \hline
\end{tabular}
\caption{Qualitative results of zero-shot MAD algorithms: EER(\%). The best-performing results are in bold.}
\label{tab:MAD_VSLLM}
\end{table*}

As a comparison to ZS-MAD using LLM, here we similarly evaluate different configurations of vision models. Based on the concept of zero-shot learning, we selected models pre-trained with related but not specific to the S-MAD task on print-scanned images. Specifically, we selected two models ResNet34 \cite{he2016deep} and VGG16 \cite{simonyan2014very} trained for the image classification task on ImageNet-1k dataset \cite{russakovsky2015imagenet}, which is commonly used for transfer learning. Regarding the distance measurement, we tested both the Euclidean distance and cosine distance to output the classification scores.  

Figure \ref{fig:MAD_VM} illustrates ZS-MAD performance using vision models with different configurations. In Figure \ref{fig:MAD_VM_lma}, the ResNet34 model shows a similar performance in detecting LMA-UBO morphs. Applying VGG16 model with Cosine distance can achieve a similar equal error rate but a higher error rate when MACER>BPCER. Together with Figure \ref{fig:MAD_VM_mipgan}, it is shown that the VGG16 model with Euclidean distance is in general not performing well on ZS-MAD task. When detecting Morph-PIPE morphs, the ResNet34 model with red and black curves is outperforming the VGG16 model as visualized in Figure \ref{fig:MAD_VM_pipe}. Overall, it is shown that the results of the VGG16 model have a higher error rate than the ResNet34 model. Employing cosine distance achieves better detection accuracy compared to Euclidean distance. 

Quantitative results are summarized in Table \ref{tab:MAD_VSLLM}. It is shown that LLM using Prompt 5 has shown the overall best performance than other prompts and vision models. Using LLM with Prompt 6 also has a considerable performance. An Obvious outperformance on Morph-PIPE morphs has been shown. As prompt 5 is explicitly asking whether the input image is bona fide, the usage of synthetic non-morphed images and morphed images has caused performance degradation when differentiating MIPGAN-II morphs.

\subsection{Comprehensive study on ZS-MAD using LLM}
\begin{figure*}[h!]
\centering
   \includegraphics[width=1\linewidth]{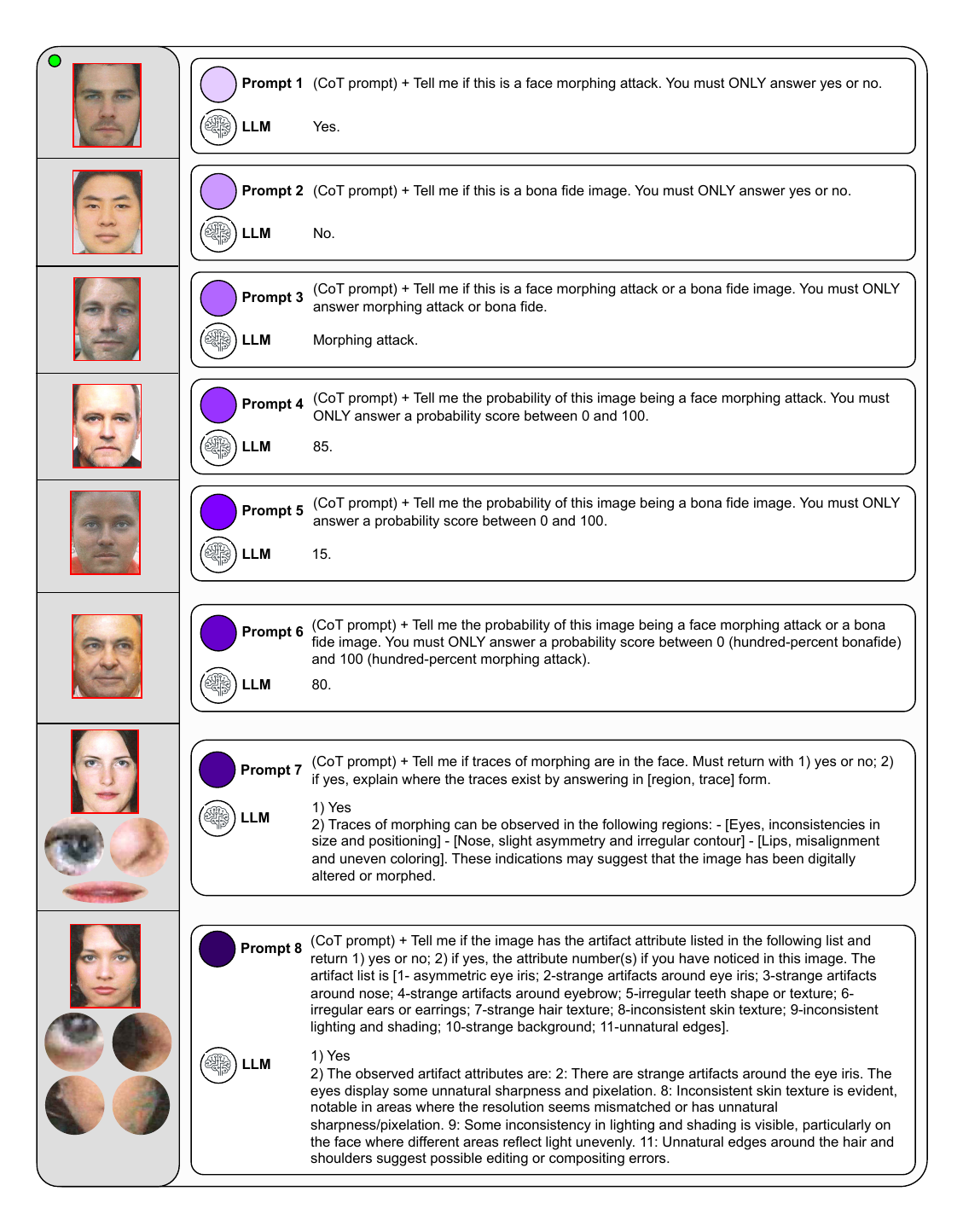}
   \caption{Examples (of correctly classified morphs) using different proposed prompts for ZS-MAD and their answers.}
\label{fig:example_prompts}
\end{figure*}

As it is an open topic for employing LLM for ZS-MAD, in the second part of the experiments, we conduct a comprehensive study of other perspectives besides the morphing attack detection accuracy.

Though we have illustrated that the results of LLM are not completely random by evaluating contrary prompts, LLM is still shown to be non-deterministic, meaning that with the same input, the model may give varied outputs \cite{song2024good}. Compared to other general applications, for the ZS-MAD task, it is more crucial as the non-determinism will directly affect the detection accuracy. In this work, we have proposed to run an evaluation by averaging scores among 5 rounds to reduce the randomness. Here we further study ZS-MAD using LLM with two perspectives: consistency of scores in different rounds and the overall stability of the method.

\begin{table}[h!]
\centering
\begin{tabular}{|c|c|ccccc|}
\hline
\multirow{2}{*}{Prompt \#} & \multirow{2}{*}{Type of morphs} & \multicolumn{5}{c|}{Fused Rounds}                                                                                    \\ \cline{3-7} 
                           &                                 & \multicolumn{1}{c|}{1}    & \multicolumn{1}{c|}{1-2}  & \multicolumn{1}{c|}{1-3}  & \multicolumn{1}{c|}{1-4}  & 1-5  \\ \hline
\multirow{3}{*}{4}         & LMA-UBO                        & \multicolumn{1}{c|}{{44}} & \multicolumn{1}{c|}{38} & \multicolumn{1}{c|}{42} & \multicolumn{1}{c|}{42} & 40  \\ \cline{2-7} 
                           & MIPGAN-II                      & \multicolumn{1}{c|}{44} & \multicolumn{1}{c|}{40}  & \multicolumn{1}{c|}{42} & \multicolumn{1}{c|}{42} & 44 \\ \cline{2-7} 
                           & Morph-PIPE                     & \multicolumn{1}{c|}{44} & \multicolumn{1}{c|}{40}  & \multicolumn{1}{c|}{42} & \multicolumn{1}{c|}{44} & 45 \\ \hline
\multirow{3}{*}{5}         & LMA-UBO                        & \multicolumn{1}{c|}{36} & \multicolumn{1}{c|}{36} & \multicolumn{1}{c|}{36} & \multicolumn{1}{c|}{40}  & 36 \\ \cline{2-7} 
                           & MIPGAN-II                      & \multicolumn{1}{c|}{36} & \multicolumn{1}{c|}{46} & \multicolumn{1}{c|}{44} & \multicolumn{1}{c|}{49} & 47 \\ \cline{2-7} 
                           & Morph-PIPE                     & \multicolumn{1}{c|}{34} & \multicolumn{1}{c|}{31} & \multicolumn{1}{c|}{34} & \multicolumn{1}{c|}{32} & 31 \\ \hline
\multirow{3}{*}{6}         & LMA-UBO                        & \multicolumn{1}{c|}{46} & \multicolumn{1}{c|}{42} & \multicolumn{1}{c|}{38} & \multicolumn{1}{c|}{38} & 37 \\ \cline{2-7} 
                           & MIPGAN-II                      & \multicolumn{1}{c|}{46} & \multicolumn{1}{c|}{41} & \multicolumn{1}{c|}{39} & \multicolumn{1}{c|}{38} & 37 \\ \cline{2-7} 
                           & Morph-PIPE                     & \multicolumn{1}{c|}{46} & \multicolumn{1}{c|}{43} & \multicolumn{1}{c|}{42} & \multicolumn{1}{c|}{42} & 41 \\ \hline
\end{tabular}
\caption{Evaluating stability of zero-shot MAD - Quantitative results when fusing different rounds: EER(\%)}
\label{tab:fuse}
\end{table}

\begin{figure}[h!]
    \centering
   \includegraphics[width=1\linewidth]{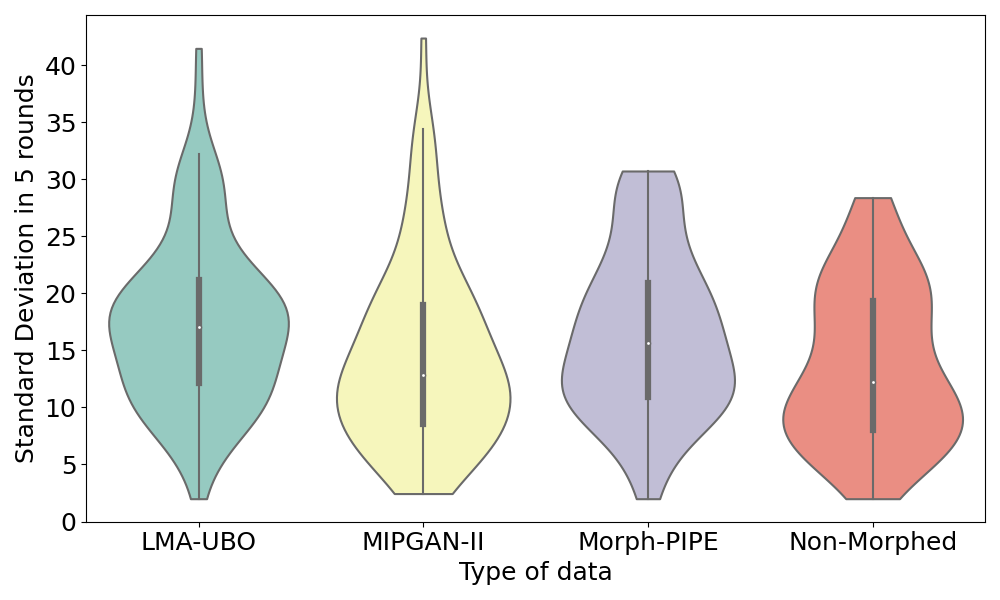}
   \caption{Stability of the ZS-MAD using LLM with Prompt 5: violin plot of the standard deviation of scores obtained in 5 rounds. Range of scores: $[0,100].$}
\label{fig:violin_prompt5}
\end{figure}

\subsubsection{Consistency and Stability}

For consistency, we select Prompt 5, where continuous scores are required for output instead of binary classes, which may lead to wider variation among different rounds for the same input. To measure the stability, the standard deviation among 5 rounds for each sample is computed and overall reported by the distribution of the value. The violin plot consisting of the inside box plots and the shape of estimated kernel density curves is shown in Figure \ref{fig:violin_prompt5}. It is illustrated that in general the same inputs are leading to consistent or similar results. It can be noticed that the two types of data which have shown a higher EER during MAD evaluation, MIPGAN-II and non-morphed data here show lower standard deviation results among different rounds.




For stability, we ran the experiments to 5 rounds and used fused scores to reduce the randomness of our results. The following question is, how does the MAD performance change and how to select a suitable number of rounds as the hyperparameter for practical applications. To address this, we measure the EER from MAD evaluation with a growing number of fusing rounds. Quantitative results are shown in Table \ref{tab:fuse}. It is shown that the MAD performances of fusing different around are in general a similar level, a few varied performances can also be noticed due to the non-determinism of LLM such as from without fusing (1) and fusing the first two rounds 1-2 using prompt 4( on) LMA or using prompt 5 on MIPGAN-II morphs. However, it is also shown that employing a fusion of 3 to 5 rounds can reasonably reduce the occasionality.

\subsubsection{Explainability}

\begin{figure*}[h!]
    \centering
   \includegraphics[width=0.8\linewidth]{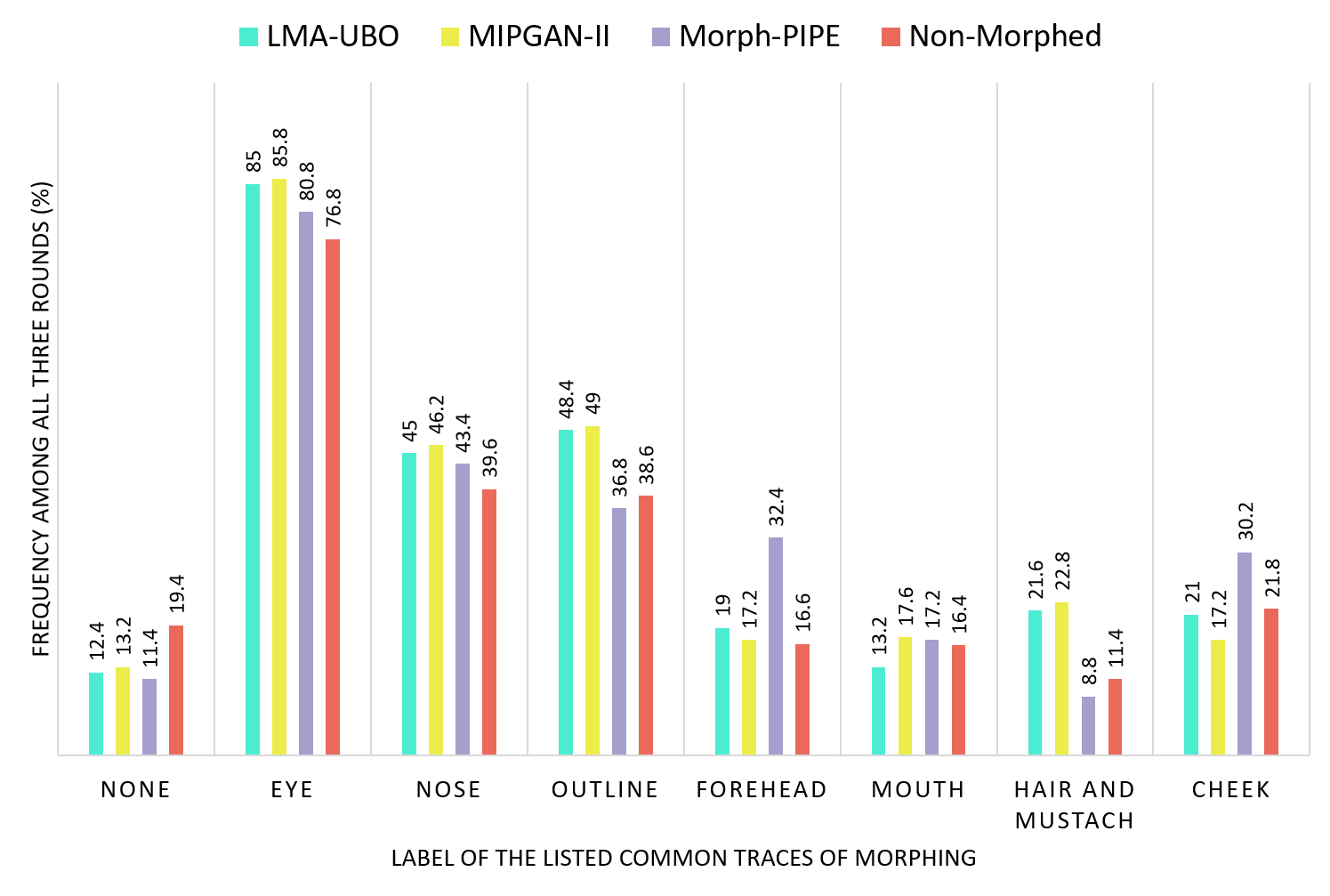}
   \caption{Detailed results of Prompt 7 with traces of morphing summarized by LLM. The frequency of the listed common traces of morphing detected by the LLM model.}
\label{fig:prompt7_item}
\end{figure*}

\begin{figure*}[h!]
    \centering
   \includegraphics[width=1\linewidth]{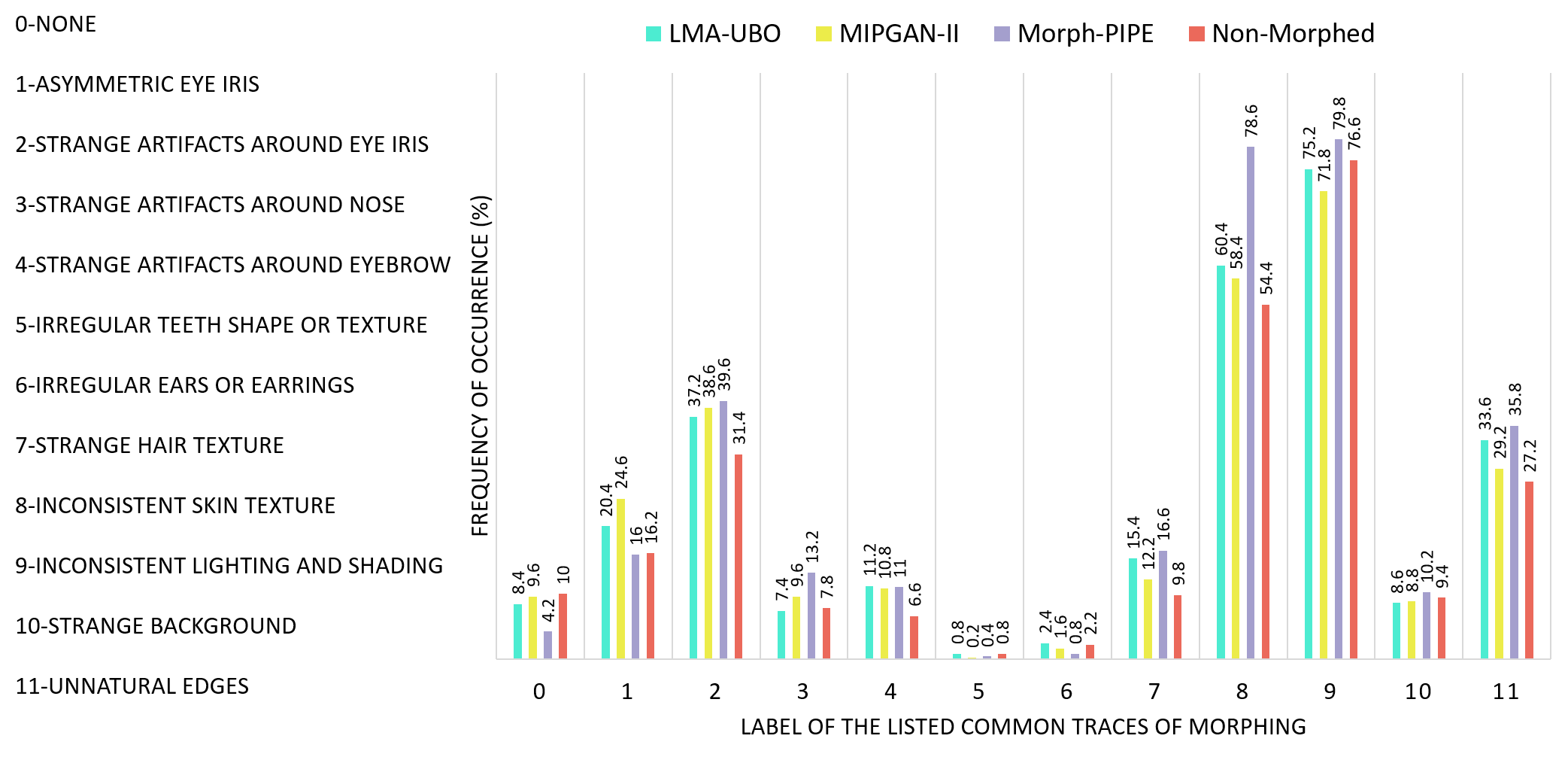}
   \caption{Detailed results of Prompt 8 with listed traces of morphing based on human prior knowledge. The frequency of the listed common traces of morphing detected by the LLM model.}
\label{fig:prompt8_item}
\end{figure*}

One of our motivations for applying LLM for MAD is the explainability to output understandable results in natural language to support end-users without a strong technical background in the real application scenario. Figure \ref{fig:example_prompts} shows examples of different proposed prompts and their corresponding answers when correctly classifying morphed images. It is shown that because we restricted the format of the answer to Prompts 1-6, the output of the model is correctly simplified as expected and can be easily extracted as binary results or scores. For Prompt 7, without any explicit guidance, the model can output explanations on detecting traces of morphing. In Prompts 7 and 8 with the reasoning of traces of morphing (either concluded by the model itself or listed manually with human knowledge), the model can also answer meaningful explanations. By referencing the answers and manually checking the image, detailed traces of morphing in the images are zoomed in circles of Figure \ref{fig:example_prompts}. This qualitatively shows the explainability of LLM for ZS-MAD. However, it should be noted that the explainability is still limited due to the non-determinism of the model.  

We further examined Prompts 7 and 8, focusing on the semantic elements describing morphing artifacts, whether these descriptions were generated by the LLM or derived from human expert knowledge. This approach not only explains the LLM’s reasoning process for the MAD task but also bridges the algorithm’s output with established human insights. Ultimately, this connection enhances the explainability of the MAD system, offering users a clearer understanding of its outcomes. Such an interpretable framework fosters greater confidence and acceptance of MAD tools in real-world scenarios.

Figure \ref{fig:prompt7_item} shows the histogram of different cases detected by LLM from different types of images. It is shown that non-morphed images are shown to have a higher frequency that no artifacts have been detected. However, similar to our observation in MAD evaluation, the results are biased for classifying images as attacks. Among the regions with artifacts detected, the eye region is reported with the highest frequency. Morph-PIPE morphs are observed to be noted with more artifacts on the forehead and cheek region while less around the region of hair and mustache. 

Similarly, the histogram for prompt 8 is visualized in Figure \ref{fig:prompt8_item}. It is shown that items 5 and 6 are rarely detected by LLM, which is consistent with the fact that our input data rarely includes ear region and wide-opened mouth and teeth because they are cropped by MTCNN model \cite{zhang2016joint} and the images are generated with face image quality assessment as control. However, it can be also observed that item 9 inconsistent lighting and shading has the highest frequency for all kinds of images including the non-morphed ones, meaning that listing of this item has caused confusion to the LLM when distinguishing between morphed and non-morphed images. Compared to the results with prompt7, many common observations can be noticed. For prompt 8, it shows that the eye region is also reported as the most common region to detect artifacts. Meanwhile, Morph-PIPE morphs are also shown to have higher frequency on skin texture. Lighting and shading appear to be a disturbing item that is detected in most of the images.


\section{Discussion}
\label{sec:discussion}
Based on our findings, the proposed zero-shot learning approach combining LLMs with general vision models proves effective for the S-MAD task. Among the tested configurations, ResNet34 paired with Cosine distance achieved the lowest EER. Additionally, the LLM configured with Prompt 5 exhibited the strongest overall detection accuracy and delivered noteworthy results when detecting Morph-PIPE morphs.

In our experiments of zero-shot MAD using LLMs, we observed that their multi-modal and text-driven nature makes them heavily dependent on obvious visual cues rather than complex image analysis. Additionally, the use of synthetic data posed limitations, particularly when attempting to differentiate MIPGAN-II morphs from non-morphed synthetic images generated by the same backbone model. Further evaluations underscore the importance of prompt engineering to effectively adapt LLMs to the S-MAD task. Simple binary classification prompts tend to introduce bias by categorizing most images as attacks. In contrast, prompts requesting probability scores reduced this bias and led to more reliable results.

In our extended evaluation of zero-shot MAD using LLMs, we observed that the proposed approach produces consistent results, and combining three to five inference rounds effectively stabilizes detection performance. Regarding explainability, our findings indicate that ZS-MAD can provide meaningful and interpretable explanations. In some cases, by visually inspecting the image, it is possible to identify actual morphing artifacts corresponding to the LLM’s descriptions. However, due to the model’s inherent non-determinism, we recommend treating these explanations as supportive guidance rather than definitive evidence. Similarly, when analyzing the semantic elements derived from prompts 7 and 8 in relation to detected traces, we found that LLMs demonstrate an understanding of the MAD task, accurately summarizing and identifying reasonable artifacts. This capability further underscores the potential of LLM-based ZS-MAD as a promising tool for practical morphing attack detection.

\section{Conclusion}
\label{sec:conclusion}
In this work, we explore zero-shot learning for S-MAD tasks to address current limitations such as insufficient training data, poor generalization, and the lack of explainability for non-technical end-users. To this end, we introduce a zero-shot approach utilizing Large Language Models (LLMs), as well as another zero-shot method based on pre-trained vision models. We design and evaluate various prompts and configurations to test their effectiveness. Our benchmark results demonstrate that both approaches can effectively detect morphing attacks without conventional training data. Moreover, we provide a detailed analysis and comprehensive evaluation, illustrating how LLM-based methods can offer meaningful, human-readable explanations, thereby making the S-MAD process more transparent and accessible.    

In this study, we restrict ourselves to a single prompt input to allow a fair comparison with ZS-MAD methods based on vision models. Future research could explore more complex prompt engineering, including step-by-step guidance or simulated real-world interactions involving human operators. Incorporating experiments with human observers might also help compare their performance against LLM-based methods, and determine whether LLM-assisted MAD can enhance human decision-making accuracy. Meanwhile, we have demonstrated the applicability of LLMs for S-MAD. Future work could investigate strategies such as fine-tuning models, incorporating few-shot samples into prompts to improve detection performance, and conducting local evaluations on non-synthetic data.

\section{Ethics Statement}

This work aims at the applicability of multi-modal large language models (LLMs) in detecting face morphing attacks. To evaluate the proposed approach, the state-of-the-art GPT-4 Turbo \cite{achiam2023gpt} API is used as the backbone model. In this context, the LLM (ChatGPT) demonstrates its potential in biometric applications by leveraging its ability to generalize in detecting unnatural visual characteristics. Additionally, its multimodal capabilities allow it to provide explainable results in natural language, making it a valuable tool to support human decision-makers. 

To comply with privacy regulations, synthetic human-identifiable data from the SynMorph synthetic face morph dataset \cite{zhang2024synmorph} were uploaded to the GPT server to generate results for this study. The synthetic face images were created using a StyleGAN2 model \cite{karras2020analyzing}, trained on the FFHQ dataset, and were then morphed using selected face morphing algorithms \cite{UBO_Morphing_Tool} \cite{zhang2021mipgan}. Since only synthetic data were utilized in this experiment, Institutional Review Board (IRB) approval was not required.

\section*{Acknowledgements}

This work was supported by the Image Manipulation Attack Resolving Solutions (iMARS) project which has received funding from the European Union’s Horizon 2020 Research and Innovation Program under Grant 883356.

\ifCLASSOPTIONcaptionsoff
  \newpage
\fi


\bibliographystyle{IEEEtran}
	\bibliography{Transactions-Bibliography/references}

\begin{thebibliography}{10}
\providecommand{\url}[1]{#1}
\csname url@samestyle\endcsname
\providecommand{\newblock}{\relax}
\providecommand{\bibinfo}[2]{#2}
\providecommand{\BIBentrySTDinterwordspacing}{\spaceskip=0pt\relax}
\providecommand{\BIBentryALTinterwordstretchfactor}{4}
\providecommand{\BIBentryALTinterwordspacing}{\spaceskip=\fontdimen2\font plus
\BIBentryALTinterwordstretchfactor\fontdimen3\font minus \fontdimen4\font\relax}
\providecommand{\BIBforeignlanguage}[2]{{%
\expandafter\ifx\csname l@#1\endcsname\relax
\typeout{** WARNING: IEEEtran.bst: No hyphenation pattern has been}%
\typeout{** loaded for the language `#1'. Using the pattern for}%
\typeout{** the default language instead.}%
\else
\language=\csname l@#1\endcsname
\fi
#2}}
\providecommand{\BIBdecl}{\relax}
\BIBdecl

\bibitem{jain2011handbook}
A.~K. Jain and S.~Z. Li, \emph{Handbook of Face Recognition}.\hskip 1em plus 0.5em minus 0.4em\relax Springer, 2011, vol.~1.

\bibitem{ngan2021face}
\BIBentryALTinterwordspacing
M.~Ngan, P.~Grother, K.~Hanaoka, and J.~Kuo, \emph{Face Analysis Technology Evaluation (FATE) Part 4: MORPH - Performance of Automated Face Morph Detection: Morph-performance of automated face morph detection}.\hskip 1em plus 0.5em minus 0.4em\relax US Department of Commerce, National Institute of Standards and Technology, 2024. [Online]. Available: \url{https://pages.nist.gov/frvt/reports/morph/frvt_morph_report.pdf}
\BIBentrySTDinterwordspacing

\bibitem{ferrara2014magic}
M.~Ferrara, A.~Franco, and D.~Maltoni, ``The magic passport,'' in \emph{IEEE International Joint Conference on Biometrics}.\hskip 1em plus 0.5em minus 0.4em\relax IEEE, 2014, pp. 1--7.

\bibitem{venkatesh2020detecting}
S.~Venkatesh, R.~Ramachandra, K.~Raja, L.~Spreeuwers, R.~Veldhuis, and C.~Busch, ``Detecting morphed face attacks using residual noise from deep multi-scale context aggregation network,'' in \emph{Proceedings of the IEEE/CVF Winter Conference on Applications of Computer Vision}, 2020, pp. 280--289.

\bibitem{UBO_Morphing_Tool}
M.~Ferrara, A.~Franco, and D.~Maltoni, ``Decoupling texture blending and shape warping in face morphing,'' in \emph{2019 International Conference of the Biometrics Special Interest Group (BIOSIG)}.\hskip 1em plus 0.5em minus 0.4em\relax IEEE, 2019, pp. 1--5.

\bibitem{PostProcessMorph}
J.~M. Singh, S.~Venkatesh, and R.~Ramachandra, ``Robust face morphing attack detection using fusion of multiple features and classification techniques,'' in \emph{2023 26th International Conference on Information Fusion (FUSION)}, 2023, pp. 1--8.

\bibitem{zhang2021mipgan}
H.~Zhang, S.~Venkatesh, R.~Ramachandra, K.~Raja, N.~Damer, and C.~Busch, ``{MIPGAN}—generating strong and high quality morphing attacks using identity prior driven {GAN},'' \emph{IEEE Transactions on Biometrics, Behavior, and Identity Science}, vol.~3, no.~3, pp. 365--383, 2021.

\bibitem{zhang2023morph}
H.~Zhang, R.~Ramachandra, K.~Raja, and C.~Busch, ``{Morph-PIPE}: Plugging in identity prior to enhance face morphing attack based on diffusion model,'' in \emph{Norsk IKT-konferanse for forskning og utdanning}, no.~3, 2023.

\bibitem{godage2022analyzing}
S.~R. Godage, F.~L{\o}v{\aa}sdal, S.~Venkatesh, K.~Raja, R.~Ramachandra, and C.~Busch, ``Analyzing human observer ability in morphing attack detection—where do we stand?'' \emph{IEEE Transactions on Technology and Society}, vol.~4, no.~2, pp. 125--145, 2022.

\bibitem{9380153}
S.~Venkatesh, R.~Ramachandra, K.~Raja, and C.~Busch, ``Face morphing attack generation and detection: A comprehensive survey,'' \emph{IEEE Transactions on Technology and Society}, vol.~2, no.~3, pp. 128--145, 2021.

\bibitem{achiam2023gpt}
J.~Achiam, S.~Adler, S.~Agarwal, L.~Ahmad, I.~Akkaya, F.~L. Aleman, D.~Almeida, J.~Altenschmidt, S.~Altman, S.~Anadkat \emph{et~al.}, ``Gpt-4 technical report,'' \emph{arXiv preprint arXiv:2303.08774}, 2023.

\bibitem{hassanpour2024chatgpt}
A.~Hassanpour, Y.~Kowsari, H.~O. Shahreza, B.~Yang, and S.~Marcel, ``{ChatGPT} and biometrics: an assessment of face recognition, gender detection, and age estimation capabilities,'' \emph{arXiv preprint arXiv:2403.02965}, 2024.

\bibitem{deandres2024good}
I.~DeAndres-Tame, R.~Tolosana, R.~Vera-Rodriguez, A.~Morales, J.~Fierrez, and J.~Ortega-Garcia, ``How good is {ChatGPT} at face biometrics? a first look into recognition, soft biometrics, and explainability,'' \emph{IEEE Access}, 2024.

\bibitem{farmanifard2024chatgpt}
P.~Farmanifard and A.~Ross, ``{ChatGPT} meets iris biometrics,'' \emph{arXiv preprint arXiv:2408.04868}, 2024.

\bibitem{chivereanu2024aligning}
R.~Chivereanu, A.~Cosma, A.~Catruna, R.~Rughinis, and E.~Radoi, ``Aligning actions and walking to {LLM-Generated} textual descriptions,'' \emph{arXiv preprint arXiv:2404.12192}, 2024.

\bibitem{venkatesh2021face}
S.~Venkatesh, R.~Ramachandra, K.~Raja, and C.~Busch, ``Face morphing attack generation and detection: A comprehensive survey,'' \emph{IEEE transactions on technology and society}, vol.~2, no.~3, pp. 128--145, 2021.

\bibitem{wu2023cheap}
G.~Wu, W.~Wu, X.~Liu, K.~Xu, T.~Wan, and W.~Wang, ``Cheap-fake detection with {LLM} using prompt engineering,'' in \emph{2023 IEEE International Conference on Multimedia and Expo Workshops (ICMEW)}.\hskip 1em plus 0.5em minus 0.4em\relax IEEE, 2023, pp. 105--109.

\bibitem{jia2024can}
S.~Jia, R.~Lyu, K.~Zhao, Y.~Chen, Z.~Yan, Y.~Ju, C.~Hu, X.~Li, B.~Wu, and S.~Lyu, ``Can {ChatGPT} detect deepfakes? a study of using multimodal large language models for media forensics,'' in \emph{Proceedings of the IEEE/CVF Conference on Computer Vision and Pattern Recognition}, 2024, pp. 4324--4333.

\bibitem{shi2024shield}
Y.~Shi, Y.~Gao, Y.~Lai, H.~Wang, J.~Feng, L.~He, J.~Wan, C.~Chen, Z.~Yu, and X.~Cao, ``{SHIELD}: An evaluation benchmark for face spoofing and forgery detection with multimodal large language models,'' \emph{arXiv preprint arXiv:2402.04178}, 2024.

\bibitem{batskos2023visualizing}
I.~Batskos, L.~Spreeuwers, and R.~Veldhuis, ``Visualizing landmark-based face morphing traces on digital images,'' \emph{Frontiers in Computer Science}, vol.~5, p. 981933, 2023.

\bibitem{singh2022fusion}
J.~M. Singh and R.~Ramachandra, ``Fusion of deep features for differential face morphing attack detection at automatic border control gates,'' in \emph{2022 10th European Workshop on Visual Information Processing (EUVIP)}.\hskip 1em plus 0.5em minus 0.4em\relax IEEE, 2022, pp. 1--5.

\bibitem{raghavendra2022multimodality}
R.~Ramachandra and G.~Li, ``Multimodality for reliable single image based face morphing attack detection,'' \emph{IEEE Access}, vol.~10, pp. 82\,418--82\,433, 2022.

\bibitem{Debiasi-PRNUVarianceMAD-BTAS-2018}
L.~Debiasi, U.~Scherhag, C.~Rathgeb, A.~Uhl, and C.~Busch, ``{PRNU} variance analysis for morphed face image detection,'' in \emph{Proc. of 9th Intl. Conf. on Biometrics: Theory, Applications and Systems ({BTAS} 2018)}, 2018.

\bibitem{borghi2023revelio}
G.~Borghi, N.~Di~Domenico, A.~Franco, M.~Ferrara, and D.~Maltoni, ``Revelio: A modular and effective framework for reproducible training and evaluation of morphing attack detectors,'' \emph{IEEE Access}, 2023.

\bibitem{zhang2024generalized}
H.~Zhang, R.~Ramachandra, K.~Raja, and C.~Busch, ``Generalized single-image-based morphing attack detection using deep representations from vision transformer,'' in \emph{Proceedings of the IEEE/CVF Conference on Computer Vision and Pattern Recognition}, 2024, pp. 1510--1518.

\bibitem{seibold2021focused}
C.~Seibold, A.~Hilsmann, and P.~Eisert, ``Focused lrp: Explainable {AI} for face morphing attack detection,'' in \emph{Proceedings of the IEEE/CVF Winter Conference on Applications of Computer Vision}, 2021, pp. 88--96.

\bibitem{myhrvold2022explainable}
H.~Myhrvold, H.~Zhang, J.~Tapia, R.~Ramachandra, and C.~G. Busch, ``Explainable visualization for morphing attack detection,'' in \emph{Proceedings of the 15th Norwegian Information Security Conference}, 2022.

\bibitem{patwardhanempowering}
S.~Patwardhan, R.~Ramachandra, and S.~Venkatesh, ``Empowering morphing attack detection using interpretable image-text foundation model,'' in \emph{9th International Conference on Computer Vision and Image Processing (CVIP)}, 2024, pp. 1--16.

\bibitem{wei2022chain}
J.~Wei, X.~Wang, D.~Schuurmans, M.~Bosma, F.~Xia, E.~Chi, Q.~V. Le, D.~Zhou \emph{et~al.}, ``Chain-of-thought prompting elicits reasoning in large language models,'' \emph{Advances in neural information processing systems}, vol.~35, pp. 24\,824--24\,837, 2022.

\bibitem{simonyan2014very}
K.~Simonyan and A.~Zisserman, ``Very deep convolutional networks for large-scale image recognition,'' \emph{arXiv preprint arXiv:1409.1556}, 2014.

\bibitem{he2016deep}
K.~He, X.~Zhang, S.~Ren, and J.~Sun, ``Deep residual learning for image recognition,'' in \emph{Proceedings of the IEEE conference on computer vision and pattern recognition}, 2016, pp. 770--778.

\bibitem{radford2021learning}
A.~Radford, J.~W. Kim, C.~Hallacy, A.~Ramesh, G.~Goh, S.~Agarwal, G.~Sastry, A.~Askell, P.~Mishkin, J.~Clark \emph{et~al.}, ``Learning transferable visual models from natural language supervision,'' in \emph{International conference on machine learning}.\hskip 1em plus 0.5em minus 0.4em\relax PMLR, 2021, pp. 8748--8763.

\bibitem{raja2017transferable}
R.~Ramachandra, K.~B. Raja, S.~Venkatesh, and C.~Busch, ``Transferable deep-{CNN} features for detecting digital and print-scanned morphed face images,'' in \emph{2017 IEEE Conference on Computer Vision and Pattern Recognition Workshops (CVPRW)}, 2017, pp. 1822--1830.

\bibitem{scherhag2020deep}
U.~Scherhag, C.~Rathgeb, J.~Merkle, and C.~Busch, ``Deep face representations for differential morphing attack detection,'' \emph{IEEE transactions on information forensics and security}, vol.~15, pp. 3625--3639, 2020.

\bibitem{zhang2024synmorph}
H.~Zhang, R.~Ramachandra, K.~Raja, and C.~Busch, ``Synmorph: Generating synthetic face morphing dataset with mated samples,'' \emph{arXiv preprint arXiv:2409.05595}, 2024.

\bibitem{karras2020analyzing}
T.~Karras, S.~Laine, M.~Aittala, J.~Hellsten, J.~Lehtinen, and T.~Aila, ``Analyzing and improving the image quality of {StyleGAN},'' in \emph{Proceedings of the IEEE/CVF conference on computer vision and pattern recognition}, 2020, pp. 8110--8119.

\bibitem{karras2019style}
T.~Karras, S.~Laine, and T.~Aila, ``A style-based generator architecture for generative adversarial networks,'' in \emph{Proceedings of the IEEE/CVF conference on computer vision and pattern recognition}, 2019, pp. 4401--4410.

\bibitem{ICAO-9303-p1-2021}
{International Civil Aviation Organization}, ``Machine readable passports -- part 1 -- introduction,'' \url{http://www.icao.int/publications/Documents/9303_p1_cons_en.pdf}, International Civil Aviation Organization (ICAO), 2021.

\bibitem{zhang2016joint}
K.~Zhang, Z.~Zhang, Z.~Li, and Y.~Qiao, ``Joint face detection and alignment using multitask cascaded convolutional networks,'' \emph{IEEE signal processing letters}, vol.~23, no.~10, pp. 1499--1503, 2016.

\bibitem{ISO-IEC-20059}
{ISO/IEC JTC1 SC37 Biometrics}, \emph{{ISO/IEC} {DIS} 20059. Information Technology -- Methodologies to evaluate the resistance of biometric recognition systems to morphing attacks}, International Organization for Standardization, 2024.

\bibitem{martin1997det}
A.~F. Martin, G.~R. Doddington, T.~Kamm, M.~Ordowski, and M.~A. Przybocki, ``The {DET} curve in assessment of detection task performance.'' in \emph{Eurospeech}, vol.~4, 1997, pp. 1895--1898.

\bibitem{russakovsky2015imagenet}
O.~Russakovsky, J.~Deng, H.~Su, J.~Krause, S.~Satheesh, S.~Ma, Z.~Huang, A.~Karpathy, A.~Khosla, M.~Bernstein \emph{et~al.}, ``Imagenet large scale visual recognition challenge,'' \emph{International journal of computer vision}, vol. 115, pp. 211--252, 2015.

\bibitem{song2024good}
Y.~Song, G.~Wang, S.~Li, and B.~Y. Lin, ``The good, the bad, and the greedy: Evaluation of {LLM}s should not ignore non-determinism,'' \emph{arXiv preprint arXiv:2407.10457}, 2024.

\end{thebibliography}
%




\end{document}